\documentclass[11pt, a4paper, logo, copyright, nonumbering]{formalmath}

\usepackage[utf8]{inputenc} 
\usepackage[T1]{fontenc}    
\usepackage{hyperref}       
\usepackage{url}            
\usepackage{booktabs}       
\usepackage{amsfonts}       
\usepackage{nicefrac}       
\usepackage{microtype}      
\usepackage{xcolor}         
\usepackage{comment}
\usepackage{fancyvrb}
\usepackage{listings}
\usepackage{algorithm}
\usepackage{algorithmicx}
\usepackage{algpseudocode}
\usepackage{subcaption}
\usepackage{cancel}
\usepackage[utf8]{inputenc}
\usepackage{amssymb}
\usepackage{fancyhdr}
\usepackage{graphicx}

\lstdefinestyle{enhanced}{
    basicstyle=\small\ttfamily,  
    backgroundcolor=\color{gray!10},
    numbers=left,
    breakindent=0pt,
    numberstyle=\tiny,
    captionpos=b,
    breakatwhitespace=false,     
    keywordstyle=\color{blue},
    commentstyle=\color{green!70!black},
    stringstyle=\color{orange},
    literate=                    
        {·}{{$\cdot$}}1
        {ℝ}{{$\mathbb{R}$}}1
        {≥}{{$\geq$}}1
        {≤}{{$\leq$}}1
        {≠}{{$\neq$}}1
        {⊢}{{$\vdash$}}1
        {✝}{{${}^\dagger$}}1
        {→}{{$\to$}}1
        {∧}{{$\land$}}1
        {∨}{{$\lor$}}1
        {¬}{{$\neg$}}1
        {∀}{{$\forall$}}1
        {∃}{{$\exists$}}1
        {λ}{{$\lambda$}}1
}
\usepackage{amsmath}
\usepackage{graphicx}       

\makeatletter
\newcommand\unmarkedfootnote[1]{%
  \begingroup
  \let\@makefntext\relax  
  \footnotetext{#1}%
  \endgroup
}
\makeatother

\title{StepFun-Prover Preview: Let's Think and Verify Step by Step}
\author[*]{
\vspace{-2ex}
{Shijie Shang}$^{1,*}$, {Ruosi Wan}$^{1,*}$, {Yue Peng}$^{1,*}$, {Yutong Wu}$^{1,2}$, {Xiong-hui Chen}$^{1,\dagger}$, {Jie Yan}$^{1,\dagger}$\newline {Xiangyu Zhang}$^{1}$ \\
\small
\vspace{-1ex}
$^1$StepFun, $^2$ University of Chinese Academy of Sciences\\
\vspace{1ex}
\small
\centering
GitHub: \url{https://github.com/stepfun-ai/StepFun-Prover-Preview},
\newline
HuggingFace: \url{https://huggingface.co/stepfun-ai/StepFun-Prover-Preview-32B}.
\vspace{-5ex}
}

\begin{document}
\begin{abstract}
We present StepFun-Prover Preview, a large language model designed for formal theorem proving through tool-integrated reasoning. Using a reinforcement learning pipeline that incorporates tool-based interactions, StepFun-Prover can achieve strong performance in generating Lean 4 proofs with minimal sampling. Our approach enables the model to emulate human-like problem-solving strategies by iteratively refining proofs based on real-time environment feedback. On the miniF2F-test benchmark, StepFun-Prover achieves a pass@1 success rate of $70.0\%$. Beyond advancing benchmark performance, we introduce an end-to-end training framework for developing tool-integrated reasoning models, offering a promising direction for automated theorem proving and Math AI assistant. 
\end{abstract}
\unmarkedfootnote{Correspondence to: Shijie Shang <shangshijie@stepfun.com>.\\Names marked with $*$ are core contributors.\\Names marked with $\dagger$ denote individuals who have departed from our team.}
\maketitle



\section{Introduction}
\label{sec:intro}

Significant progress has been made in automated theorem provers recently, primarily driven by the emergence and deployment of large language models (LLMs) with long chain-of-thought ~\cite{jaech2024openai,dsr1_cite,qwq2024preview}. Recent breakthroughs~\cite{kimina_prover_2025,ren2025deepseekproverv2advancingformalmathematical,lin2025goedelproverfrontiermodelopensource,zhou2025solvingformalmathproblems} effectively employs LLMs' reasoning capabilities for formal theorem proving through designing sophisticated informal thinking patterns, or refined environmental interaction strategy to derive complex formal proofs.

\begin{figure}[htpb]
    \centering
    \includegraphics[width=0.7\linewidth]{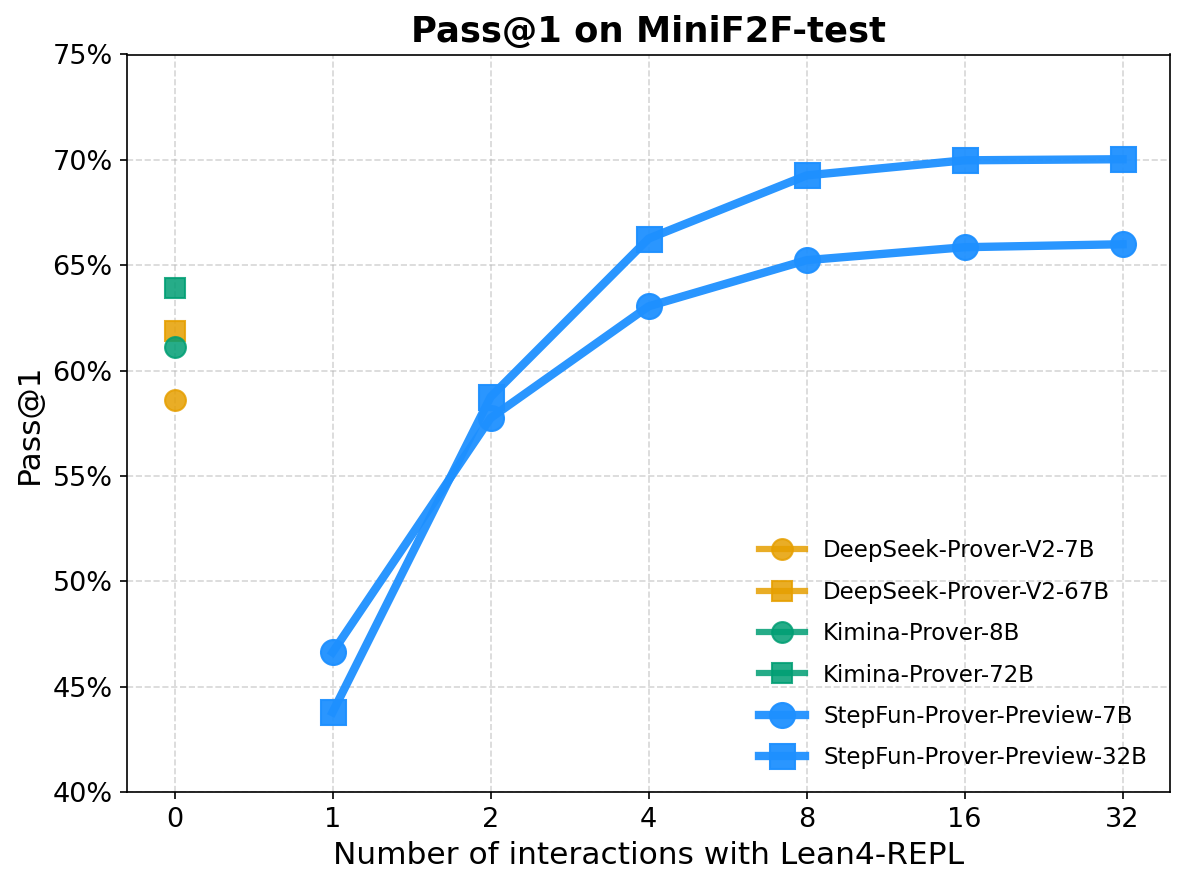}
    \caption{Performance comparison on MiniF2F-test. y-axis shows the pass@1, which is computed by averaging 32 trials; while x-axis denotes the maximum number the provers are allowed to interact with Lean4-REPL before getting successful proof. Note both DeepSeek-Prover and Kimina-Prover utilize at least 32K token context length. Stepfun-Prover was evaluated using 20K context window including feedback from Lean4-REPL.}
    \label{fig:1}
\end{figure}


Although the aforementioned methods are effective, they typically require substantially more attempts to achieve satisfactory performance. According to published results~\cite{Kimina_prover_2025_plus,ren2025deepseekproverv2advancingformalmathematical}, scaling proof sampling from $32$ to $64$ attempts yields marginal accuracy gains of merely $3\% \sim 4\%$, indicating diminishing returns. This inefficiency suggests possibly persistent error patterns cause consistent failures across multiple attempts for the same problem. The substantial number of proof attempts required by formal theorem provers to reach correct solutions poses a significant usability barrier, limiting their adoption by non-expert users. In contrast, skilled Lean 4~\cite{moura2021lean} users can complete a proof within a few attempts, demonstrating significantly greater efficiency than the formal theorem prover. The fundamental limitation lies in formal theorem provers' inability to fully leverage Lean 4 feedback due to their rigid, human-predefined reasoning patterns, which necessitates excessive sampling attempts during proof search. Meanwhile, skilled Lean 4 users usually exhibit efficient error diagnosis by interpreting feedback, enabling adaptive strategy selection for error resolution. Then a question naturally rises: 

\begin{center}
\textit{Can formal theorem provers self-evolve suitable patterns to think, interact with formal verifier, and derive the formal proof efficiently?}
\end{center}


In this work, we aim to explore this question by allowing the model to autonomously decide when to stop using tools, without constraining the number of Lean 4 interactions. By doing so, we encourage the model to engage in open-ended, tool-assisted reasoning, driven entirely by its own perception of failure and feedback—much like how humans learn and correct themselves through trial, verification, and reflection. Our primary contributions are: 
\begin{enumerate}
    \item \textbf{Tool-integrated reasoning for formal theorem prover.} We propose a dynamic reasoning framework that enables LLMs to agentically control interactions with the Lean 4 environment. The process interleaves natural language reasoning, Lean 4 code snippets, and feedback from the real-time environment prior to the final generation of the answers.
    \item \textbf{Tool-integrated reinforcement learning pipeline with cold start.} Our approach develops tool-integrated reasoning through sequential training phases: initial cold-start training establishes fundamental environment interaction patterns, followed by iterative refinement through reinforcement learning from proof outcomes combined with supervised fine-tuning for precise capability improvement.
\end{enumerate}

The experimental results from Figure~\ref{fig:1} show that our trained model, \textbf{StepFun-Prover-Preview-7B}, can catch up DeepSeek-Prover-V2-671B and Kimina-Prover-72B on $pass@1$, while \textbf{StepFun-Prover-Preview-32B} surpasses all its counterparts by over $4\%$ on $pass@1$. We will open source StepFun-Prover-Preview-7B and StepFun-Prover-Preview-32B soon.

\section{Methodology}
\label{sec:method}
\begin{figure}[h]
  \centering
  \includegraphics[width=0.78\linewidth]{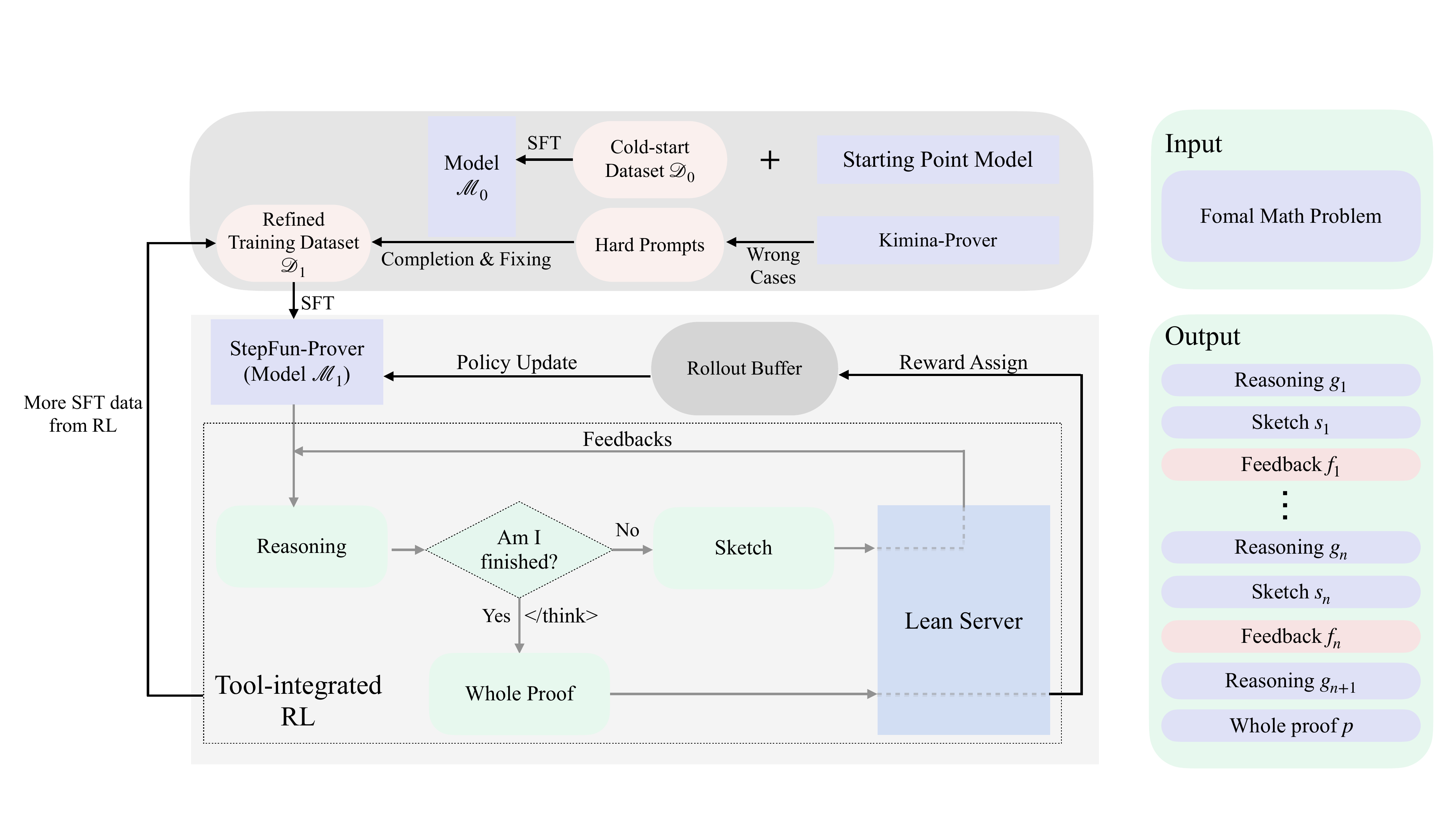}
  \vspace{-2ex}
  \caption{Our training pipeline. Left. Tool-integrated RL and iterative RL-SFT cycle. The upper left illustrates data preparation. Right. Tool-integrated reasoning pattern.
  }
  \label{fig:algorithm-pipeline}
  \vspace{-2ex}
\end{figure}

In this section, we present our end-to-end training pipeline (see Figure ~\ref{fig:algorithm-pipeline}) for developing tool-integrated reasoning models.

\vspace{-2ex}

\subsection{Cold Start for Tool-integrated Reasoning}

We first compile existing open-sourced formal mathematics datasets from multiple sources, including the Lean Workbook~\cite{LWB} and the STP dataset ~\cite{stp}. Additionally, we employ Kimina-Autoformalizer-7B ~\cite{kimina_prover_2025} to autoformalize selected problems from the Numina Math 1.5 dataset. However, our analysis reveals that both the open-sourced formal problems and those generated via autoformalization exhibit several recurring issues, such as:

\begin{itemize}
\item \textbf{Tautological statements}, where the conclusion trivially follows from the premises without meaningful reasoning.
\item \textbf{Contradictory premises}, rendering the problem unsolvable or logically inconsistent.
\item \textbf{Irrelevant conclusions}, where the statement to prove does not meaningfully relate to the given premises.
\item \textbf{Trivial or ill-posed problems}, which lack mathematical depth or are poorly structured.
\end{itemize}

To ensure data quality, we employ DeepSeek-V3 ~\cite{dsv3_cite} to filter out low-quality problems. Subsequently, we construct a cold start dataset $\mathcal{D}_{0}$ through two parallel processes:

\begin{itemize}
    \item \textbf{Multi-turn data collection. } We use Claude Sonnet 4 ~\cite{Claude4} to gather approximately 8,000 effective multi-turn trajectories, which serve as the initial multi-turn reasoning data.
    \item \textbf{Single-turn data processing. } We refine the proofs generated by Kimina-Prover-72B ~\cite{Kimina_prover_2025_plus} by removing all intermediate tactic blocks and reasoning steps between them, retaining the natural language reasoning of the problem and the final answer in Lean 4. This modification is motivated by our observation that, in some cases, the final answer deviates from the logical flow of the preceding tactic blocks, suggesting that intermediate reasoning may not be necessary.
\end{itemize}

\subsection{Two-stage Supervised Fine-tuning}
We adopt the following two-stage training approach:
\begin{enumerate}
    \item  We first supervised fine-tune the starting point model using question-answer pairs from the open-sourced STP dataset~\cite{stp}. This phase establishes the model's basic understanding of Lean 4 syntax and proof structure.
    \item Subsequently, we perform continual supervised fine-tuning using our cold start dataset $\mathcal{D}_{0}$ to establish initial tool-integrated reasoning capabilities and proof generation capabilities. This two-stage process yields our cold-started model, denoted as $\mathcal{M}_{0}$.
\end{enumerate}

\subsection{Response Patterns Fusion}
While the cold-started model $\mathcal{M}_{0}$ demonstrated basic environment interaction capabilities, its performance fell significantly below expectations. Our analysis of the response generated by the model $\mathcal{M}_{0}$ revealed a critical limitation: $\mathcal{M}_{0}$ exhibited difficulty in effectively integrating the distinct response patterns from Claude and Kimina styles. 

To address this challenge, we developed a novel fusion approach:

\begin{enumerate}
    \item \textbf{Wrong responses collection.} We first gathered incorrect responses generated by Kimina-Prover-72B.
    \item \textbf{Correction and dataset construction.} Using $\mathcal{M}_{0}$, we systematically corrected these responses while preserving valid reasoning trajectories. This process yielded our refined training dataset $\mathcal{D}_1$.
    \item \textbf{Model enhancement.} We performed supervised fine-tuning using exclusively $\mathcal{D}_1$ based on the starting point model, resulting in our improved model $\mathcal{M}_1$. This targeted training approach specifically enhanced the model's ability to harmonize the different response styles. The model $\mathcal{M}_1$ serves as the starting point for subsequent reinforcement learning.
\end{enumerate}

\subsection{Tool-integrated Reinforcement Learning}

During reinforcement learning stage, our rollout process differs from the conventional approach, which typically generates only text-based reasoning. By contrast, our rollout approach integrates the collaboration of a LLM with a sandbox-style environment, enabling the production of hybrid content that combines text, Lean 4 code snippets enclosed by $\texttt{<sketch>}$ and $\texttt{</sketch>}$, and real-time environment feedback enclosed by $\texttt{<REPL>}$ and $\texttt{</REPL>}$. Concretely, the generation starts with text-based reasoning $g_1$. Then the model propose the first Lean 4 solution $s_1$ to the problem. When a sketch termination trigger ($\texttt{</sketch>}$) is detected, the generation pause and the Lean 4 solution $s_1$ is parsed and send to sandbox environment for execution. Upon completion, the sandbox’s output $f_1$ (successful results or error messages) is filled within $\texttt{<REPL></REPL>}$ tags and feedback to the model, which continues generating the rollout until either providing a final answer $p$ or producing a new Lean 4 solution, ultimately producing a hybrid reasoning trajectory $[g_1 \oplus s_1 \oplus f_1 \oplus\cdots\oplus p]$.

When the model believes it has reached a final answer, it is required to generate a special delimiter $\texttt{</think>}$ and place its final Lean 4 solution after this marker, enabling reliable rule-based verification. We evaluate the correctness of the code segment following $\texttt{</think>}$ to assign the trajectory reward: if the proof passes REPL verification, the reward is set to $1$; otherwise, $0$. This binary outcome-based reward design simplifies supervision, mitigates reward hacking, and promotes more diverse problem-solving behavior by focusing only on the final result~\cite{dsr1_cite}. We perform reinforcement learning from the instruction-tuned model $\mathcal{M}_1$ using GRPO, based on the tool-integrated rollout and reward setup. In line with ~\cite{hu2025openreasonerzeroopensourceapproach}, which applies PPO without KL regularization, we similarly omit the KL term in our GRPO framework to promote exploration from a well-aligned instruct model. Given a training set $\mathcal{Q}$, for each sample $q \in \mathcal{Q}$, we sample $G$ candidate answers ${o_1, o_2, \dots, o_G}$ from the old policy $\pi_{\theta_{\text{old}}}$. Each output $o_i$ is evaluated to obtain its corresponding binary reward $r_i$, resulting in a reward group ${r_1, r_2, \dots, r_G}$. The group-based advantage for the $i$-th sample is then computed as:

\begin{equation}
    A_i = \frac{r_i - \text{mean}({r_1, r_2, \dots, r_G})}{\text{std}({r_1, r_2, \dots, r_G})}
\end{equation}

The objective is then defined as:
\begin{align}
    \mathcal{J}(\theta) &= \mathbb{E}_{q \sim \mathcal{Q}, \{o_i\}_{i=1}^G \sim \pi_{\theta_{\text{old}}}(O|q)} \left[ \frac{1}{G} \sum_{i=1}^{G} \left( \min \left( \frac{\pi_\theta(o_i|q)}{\pi_{\theta_{\text{old}}}(o_i|q)} A_i, \, \text{clip} \left( \frac{\pi_\theta(o_i|q)}{\pi_{\theta_{\text{old}}}(o_i|q)}, 1 - \varepsilon, 1 + \varepsilon \right) A_i \right) \right) \right]
\end{align}

\begin{figure}[htbp]
    \centering
    \includegraphics[width=0.8\textwidth]{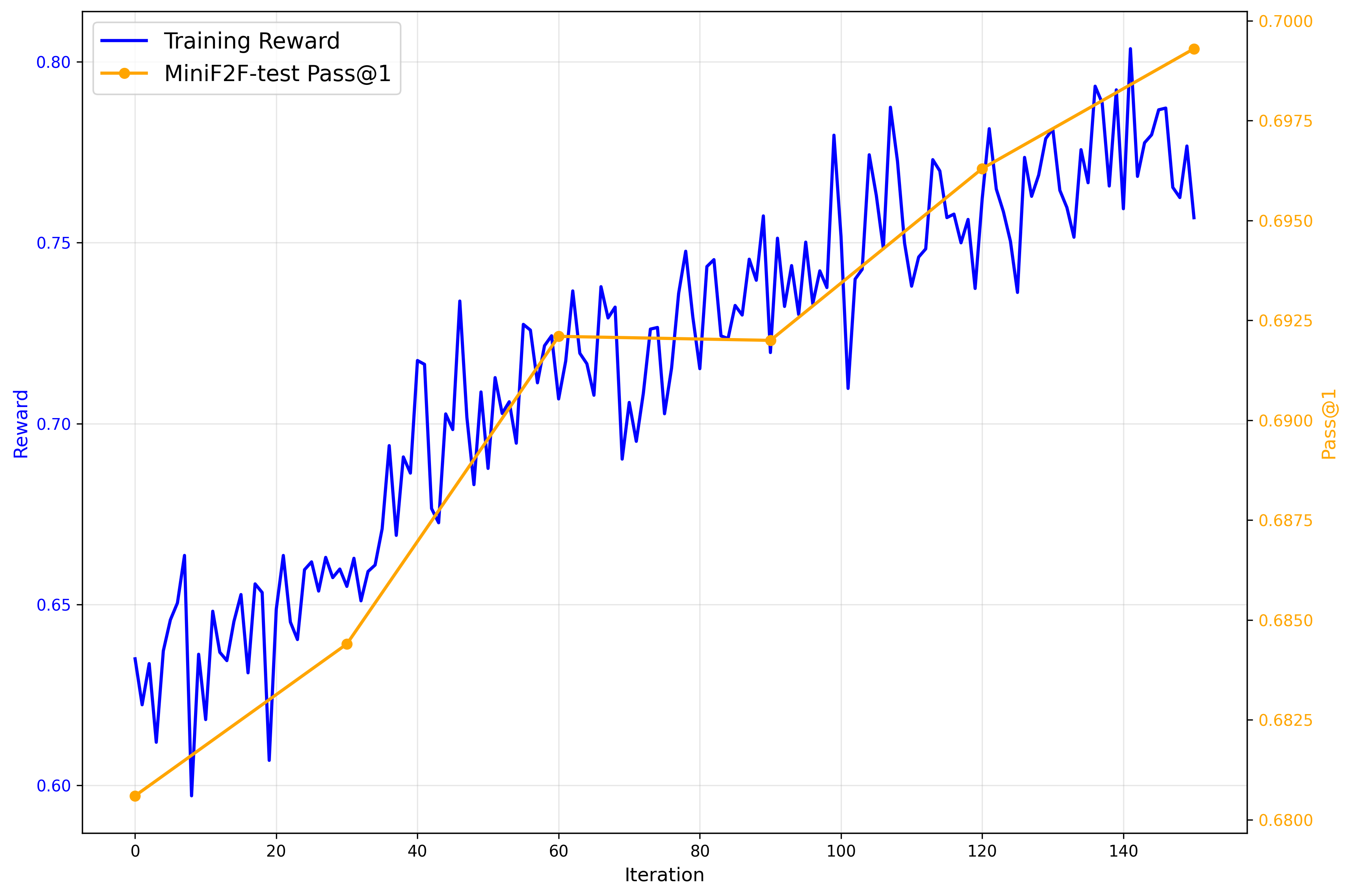}
    \caption{Training reward and test accuracy (pass@1) of the tool-integrated GRPO.}
    \label{fig:RL}
\end{figure}

\subsection{Iterative RL-SFT Cycle}

Upon convergence of the tool-integrated reinforcement learning, we collect correct trajectories from problems scoring below 0.5 accuracy during training. These trajectories are then filtered to retain only those demonstrating comprehensive REPL feedback analysis. The selected trajectories subsequently undergo another round of supervised fine-tuning. Finally, we implement a secondary reinforcement learning phase to further enhance the model's tool-integrated reasoning capabilities.

\subsection{Training Details}
\subsubsection{Supervised Fine-tuning}
We train our 7B and 32B models based on DeepSeek-R1-Distill-Qwen-7B and DeepSeek-R1-Distill-Qwen-32B \cite{dsr1_cite} respectively. We implement distinct supervised fine-tuning (SFT) strategies for non-reasoning data and tool-integrated reasoning data.

\textbf{SFT of non-reasoning data.} We train the model on code completion data using a multi-step learning rate schedule with a 10\% warmup phase. Training is conducted over two epochs: the learning rate is linearly increased to a maximum of 2e-5 during the warmup, held steady, and then decayed by a factor of 10 in the second epoch. To maintain compatibility with the response patterns of the base model, we preserve the use of the special <think></think> token structure in the training data, but intentionally leave the thinking content blank in this stage.

\textbf{SFT of tool-integrated reasoning data.} We introduce training data that includes explicit reasoning patterns, model interpretation of REPL feedback, and the corresponding follow-up actions based on such interpretation. This encourages the model to iteratively refine its reasoning in response to environment signals. We adopt a linear learning rate schedule with a 10\% warmup phase, and train for three epochs. Besides, we apply a weight decay of 1e-2 to prevent overfitting. Additionally, we mask out the $\texttt{<REPL></REPL>}$ feedback during loss computation. This sandbox-based output masking approach blocks external tokens from interfering with loss calculations, ensuring training stability and preserving the model’s inherently generated coherent reasoning sequences from disruption.

\subsubsection{Reinforcement Learning}

\textbf{RL parameter configuration.} We set the rollout batch size to $512$ and generate $8$ samples per problem. The maximum generation length is set to 16384 tokens, and the REPL execution timeout is set to 60 seconds. During rollout, we use a sampling temperature of 1.0 and a top-$p$ value of 0.999, encouraging exploration. 

\textbf{RL training data curation.} Our training prompts are derived from the STP dataset~\cite{stp}. To prevent the model from exploiting memorized SFT examples as shortcuts, we first remove all prompts used during the SFT stage from the candidate pool. Instead of filtering data statically before training, we integrate data selection into the RL training loop itself. Specifically, we extract only the single-turn subset from the second-stage SFT dataset $\mathcal{D}_0$ and use it to train a 7B model, denoted as $\mathcal{M}_1’$. We then use $\mathcal{M}_1’$ as the initialization for a single-turn RL process focused on data filtering and task selection, with 32 sampled outputs per prompt. During each RL training step, we collect all rollouts generated by the model and use them to continuously update the policy. We aggregate all generated data across the entire RL process and compute a per-prompt success rate based on REPL feedback. We retain only those prompts whose success rates fall strictly between 0 and 1. This effectively removes problems that are either consistently trivial or currently unsolvable given the model’s capabilities. Compared to the conventional approach of filtering with a static SFT model before training, our dynamic, in-training filtering strategy is better suited to identifying high-value prompts—those that may be too difficult for the initial model, but become solvable as the model improves. This allows us to focus training on problems that are most likely to drive meaningful capability gains.

\textbf{Lean 4 server.} We wrapped Lean4-REPL~\cite{lean4repl} by python to enable multi-turn interaction between Lean 4 and LLM during RL training. In order to enhance the efficiency of the interaction and to maximize GPU utilization during training, we have taken the following measures:

\begin{itemize}
    \item Modified the main function of Lean4-REPL to optimize memory reclamation efficiency, reducing the frequency of memory explosions caused by repeated REPL process invocations and minimizing REPL restarts\footnote{Our modification on Lean4-REPL have been acknowledged by the official developers. More details can be seen in \url{https://github.com/leanprover-community/repl/pull/83}.}.
    \item Analyzed and categorized all possible return types from the REPL, implementing comprehensive handling for each scenario to enhance stability in large model-REPL interactions. Notably, we set a 60-second timeout threshold to prevent model misuse of search-based tactic proof-search and accelerate Lean4-REPL efficiency;
    \item Implemented asynchronous mechanism for both multi-turn input streaming of LLM, and submission of generated proof drafts to Lean4-REPL servers, eliminating delays caused by variations in generation time, proof step count, or Lean4-repl verification time across samples;
    \item Established a Lean4-REPL Remote Server using Redis~\cite{redis} as a central message broker based on a Producer-Broker-Consumer pattern, enabling concurrent execution of over 1,000 Lean4-REPL processes for multi-turn training. Stability-focused design isolates large model training from Remote Server instability.
\end{itemize}

These Lean4-REPL remote server optimizations accelerated our multi-turn RL training by over $10x$ faster, and enabled feasible multi-turn RL training with unlimited interaction times.
\section{Experimental Results}
\label{sec:exps}
All experimental results of StepFun-Prover-Preview are conducted with Lean 4.20.0. Without further specification, baseline evaluation results are sourced from their respective original papers. For all our models, the maximum generation length is set to $20480$ tokens unless otherwise specified. We set the REPL timeout to $60$ seconds. We use a temperature of $1.0$, a top-p value of $0.999$, and generate $32$ responses per query to estimate pass@1. 

\subsection{Results on MiniF2F Benchmark}
We evaluate on Numina's corrected miniF2F benchmark from their HuggingFace repository. StepFun-Prover-Preview-32B achieves $70.0\%$ pass@1, surpassing prior SOTA results with a smaller model size, as shown in Table~\ref{table_compare_sota}. This demonstrates the effectiveness of our tool-integrated reasoning approach in low-pass settings.

\begin{table}[htbp]
  \centering
  \caption{Performance on MiniF2F-test}
  \label{table_compare_sota}
  \begin{tabular}{lcrc}  
    \toprule
    \textbf{Model} & \textbf{Pass@1}  \\
    \midrule
    DeepSeek-Prover-V2-7B & $58.6\%$   \\
    DeepSeek-Prover-V2-671B & $61.9\%$   \\
    Kimina-Prover-8B & $61.1\%$   \\
    Kimina-Prover-72B & $63.9\%$   \\
    StepFun-Prover-Preview-7B & $66.0\%$   \\ 
    StepFun-Prover-Preview-32B & $70.0\%$  \\
    \bottomrule
  \end{tabular}
\end{table}

\subsection{Performance Analysis}
\subsubsection{Test Time Scaling in Tool-integrated Reasoning}
Tool-integrated reasoning models scale performance with test-time compute. As shown in Table~\ref{generation_length_scaling}, StepFun-Prover-Preview-32B shows consistent performance improvements on the miniF2F-test benchmark when we extend the maximum generation length.

\begin{table}[htbp]
  \centering
  \caption{Performance of StepFun-Prover-Preview-32B on MiniF2F-test}
  \label{generation_length_scaling}
  \begin{tabular}{ccrc}  
    \toprule
    \textbf{Maximum generation length} & \textbf{Pass@1}  \\
    \midrule
    $4096$ & $58.3\%$   \\
    $8192$ & $66.5\%$   \\
    $12288$ & $68.9\%$   \\
    $16384$ & $69.9\%$   \\
    $20480$ & $70.0\%$ \\
    \bottomrule
  \end{tabular}
\end{table}

\begin{figure}[htbp]
    \centering
    \includegraphics[width=0.8\textwidth]{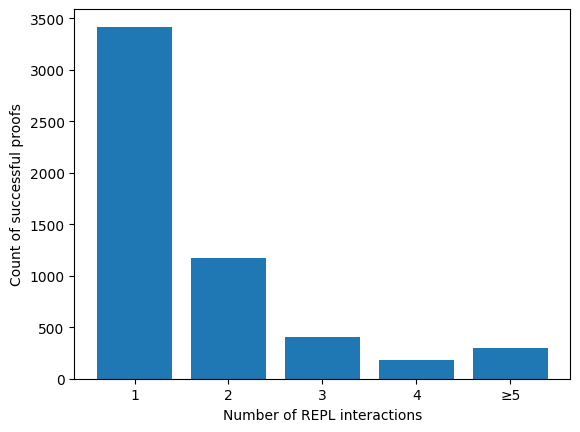}
    \caption{REPL interaction frequency distribution}
    \label{fig:repl_hist}
\end{figure}

As shown in Figure \ref{fig:repl_hist}, the REPL interaction distribution for successful proofs on the miniF2F-test benchmark (using StepFun-Prover-Preview-32B with 32 responses per problem) reveals the robust ability of StepFun-Prover-Preview to solve problems through extended interaction sequences during reasoning process, demonstrating that correct solutions can emerge after multiple rounds of refinement.

\subsubsection{Diverse Reasoning Patterns}
Through our tool-integrated reinforcement learning framework with cold start, StepFun-Prover-Preview develops several advanced reasoning capabilities that leverage real-time environment feedback, including but not limited to:
\begin{itemize}
    \item Proactive resolution of Lean 4 warnings in the REPL  even when no errors are present
    \item Dynamic proof restructuring when execution exceeds time thresholds (e.g., 60 seconds)
    \item Iterative solution improvement through REPL feedback analysis (see Appendix \ref{sec:examples})
\end{itemize}

\section{Related Work}
\label{sec:RelatedWork}

Prior to the advent of reasoning models~\cite{jaech2024openai, openai2025o3o4mini, dsr1_cite, qwq2024preview, qwq2025}, formal theorem provers primarily employed LLMs either to generate individual proof steps ~\cite{GuHyperTree, PoluFormalMath,alphaproof2024,InternProver,BFSProver,DeepSeekProverV1p5} or to directly produce complete proofs \cite{DeepSeekProverV1p5,lin2025goedelproverfrontiermodelopensource,Baldur,DSP,LEGO}. To improve proof search exploration, LLMs-based approaches often combine with traditional tree search methods, such as Best-First Search ~\cite{PoluFormalMath,InternProver,BFSProver}  and Monte Carlo Tree Search \cite{GuHyperTree, alphaproof2024, DeepSeekProverV1p5}. Central to these systems, the search algorithm navigates proof paths via heuristics, offering improved exploration at the expense of increased computational overhead.

Recent advanced reasoning models, such as OpenAI o-seriers~\cite{jaech2024openai, openai2025o3o4mini}, DeepSeek R1~\cite{dsr1_cite}, and QwQ~\cite{qwq2024preview, qwq2025}, have significantly improved the reasoning capabilities of large language models (LLMs). These models exhibit a human-like ability to reflect on and revise their reasoning upon recognizing mistakes, a paradigm shift that has substantially advanced model performance in domains such as mathematics solving~\cite{openai2025o3o4mini, comanici2025gemini} and code generation~\cite{el2025competitive, openai2025codex, comanici2025gemini}. Early reasoning models typically used feedback from external environments only as a final correctness check~\cite{dsr1_cite, lin2025goedelproverfrontiermodelopensource, ren2025deepseekproverv2advancingformalmathematical}, without incorporating it as an additional signal during the intermediate reasoning process. More recent work~\cite{xue2025simpletir,AgenticReasoning,Nemotron,ReTool,Torl} begins to explore multi-turn, tool-integrated reasoning, enabling models to interact iteratively with environments such as REPLs, search engines, or formal verifiers. This paradigm not only improves factual accuracy and sample efficiency, but also more closely mirrors human approaches to complex problem solving. 

In formal mathematics, several recent efforts~\cite{kimina_prover_2025, ren2025deepseekproverv2advancingformalmathematical, lin2025goedelproverv2, zhou2025solvingformalmathproblems,LPV2} have achieved impressive results on benchmarks like miniF2F by allowing models to revise faulty reasoning and incrementally improve their proofs through interaction. However, while effective, these designs often abstract away the learning dynamics and limit the model’s autonomy. In particular, they overlook a key opportunity: leveraging environment feedback alone at test time, \textbf{without additional supervision}, to support self-directed tool use, iterative reflection, and adaptive exploration—capabilities that are essential in settings like formal mathematics, where feedback is sparse but highly reliable.
\section{Conclusion}
\label{sec:conclusions}

In this work, we introduce an end-to-end training pipeline for developing a tool-integrated reasoning model, featuring cold-start data synthesis and tool-integrated reinforcement learning. The resulting model, StepFun-Prover-Preview, achieves significant accuracy improvements with minimal sampling ($70.0\%$ pass@1 on miniF2F-test) by leveraging feedback from the real-time environment for iterative proof refinement. This demonstrates the efficacy of combining human-like reflection with formal verification. Our future work will focus on transferring formal reasoning rigor to enhance the reasoning ability of LLMs, and developing multi-agent systems for advanced mathematical discovery.


\section{Acknowledgements}
This work was supported by computing resources and infrastructure provided by \href{https://www.stepfun.com/}{StepFun}. We are grateful to our StepFun colleagues for their valuable feedback and contributions.

\newpage
\bibliographystyle{unsrt}
\bibliography{references.bib}
\newpage

\appendix
\section{Examples of Tool-integrated Reasoning}
\label{sec:examples}
System prompt:
\begin{lstlisting}[style=enhanced]
You will be given an unsolved Lean 4 problem. Think carefully and work towards a solution. At any point, you may use the Lean 4 REPL to check your progress by enclosing your partial solution between <sketch> and </sketch>. The REPL feedback will be provided between <REPL> and </REPL>. Continue this process as needed until you arrive at a complete and correct solution.
\end{lstlisting}
User input:
\begin{lstlisting}[style=enhanced]
```lean4
import Mathlib

theorem test_theorem (x y z : ℝ) (hx : 0 < x) (hy : 0 < y) (hz : 0 < z) :
    (x^2 - z^2) / (y + z) + (y^2 - x^2) / (z + x) + (z^2 - y^2) / (x + y) ≥ 0 := by
```
\end{lstlisting}
Output:
\begin{lstlisting}[style=enhanced]
I'll tackle this algebraic inequality step by step, formalizing the proof in Lean 4. The inequality is:

$$(x^2 - z^2) / (y + z) + (y^2 - x^2) / (z + x) + (z^2 - y^2) / (x + y) \geq 0$$

for positive real numbers $x, y, z$.

First, let's think about how to approach this algebraically. One strategy is to find a common denominator and analyze the numerator. The common denominator would be $(y+z)(z+x)(x+y)$. Let's compute the numerator:

For $(x^2 - z^2) / (y + z)$: Multiply by $(z+x)(x+y)$ to get $(x^2 - z^2)(z+x)(x+y)$
For $(y^2 - x^2) / (z + x)$: Multiply by $(y+z)(x+y)$ to get $(y^2 - x^2)(y+z)(x+y)$
For $(z^2 - y^2) / (x + y)$: Multiply by $(y+z)(z+x)$ to get $(z^2 - y^2)(y+z)(z+x)$
So the numerator becomes: $(x^2 - z^2)(z+x)(x+y) + (y^2 - x^2)(y+z)(x+y) + (z^2 - y^2)(y+z)(z+x)$

Let's expand each term:

$(x^2 - z^2)(z+x)(x+y) = (x^2 - z^2)(zx + z^2 + x^2 + xy)$
$(y^2 - x^2)(y+z)(x+y) = (y^2 - x^2)(xy + y^2 + zx + z^2)$
$(z^2 - y^2)(y+z)(z+x) = (z^2 - y^2)(yz + z^2 + xy + xz)$
This looks complicated, but maybe we can find a pattern or factorization. Let's try a different approach. Notice that each term is of the form $(a^2 - b^2)/(c + d)$. We can use the identity $a^2 - b^2 = (a-b)(a+b)$:

$(x^2 - z^2)/(y+z) = (x-z)(x+z)/(y+z)$
$(y^2 - x^2)/(z+x) = (y-x)(y+x)/(z+x)$
$(z^2 - y^2)/(x+y) = (z-y)(z+y)/(x+y)$
So our expression becomes: $(x-z)(x+z)/(y+z) + (y-x)(y+x)/(z+x) + (z-y)(z+y)/(x+y)$

This is still complex, but perhaps we can find a way to show it's non-negative. One approach is to use the AM-GM inequality or other algebraic manipulation.

Let's try to rearrange the expression. Notice that:

$(x-z)(x+z)/(y+z) = (x^2 - z^2)/(y+z)$
$(y-x)(y+x)/(z+x) = (y^2 - x^2)/(z+x)$
$(z-y)(z+y)/(x+y) = (z^2 - y^2)/(x+y)$
If we can show that each term is non-negative, or that their sum is non-negative, we'd be done. However, since we have positive variables, we need to be careful with the signs.

Wait, let's think about this differently. Consider the function $f(a,b) = (a^2 - b^2)/(c + d)$ where $c$ and $d$ are positive. If we can show that the sum of these terms is non-negative, we'd be done.

Alternatively, we can try to find a common denominator and analyze the numerator. Let's compute the numerator explicitly:

$(x^2 - z^2)(z+x)(x+y) + (y^2 - x^2)(y+z)(x+y) + (z^2 - y^2)(y+z)(z+x)$

Expanding each term:

$(x^2 - z^2)(z+x)(x+y) = (x^2 - z^2)(zx + z^2 + x^2 + xy)$
$(y^2 - x^2)(y+z)(x+y) = (y^2 - x^2)(xy + y^2 + zx + z^2)$
$(z^2 - y^2)(y+z)(z+x) = (z^2 - y^2)(yz + z^2 + xy + xz)$
This is quite complex, but perhaps we can find a pattern. Let's try to compute the full expansion:

First term: $(x^2 - z^2)(z+x)(x+y) = (x^2 - z^2)(zx + z^2 + x^2 + xy)$ $= x^2(zx + z^2 + x^2 + xy) - z^2(zx + z^2 + x^2 + xy)$ $= x^3z + x^2z^2 + x^4 + x^3y - z^3x - z^4 - z^2x^2 - z^3y$

Similarly for the other terms. After expanding all three terms and combining like terms, we'd get a polynomial in $x, y, z$.

However, instead of expanding everything, let's try a different approach. Notice that the expression is cyclic in $x, y, z$. Maybe we can assume an order (e.g., $x \geq y \geq z$) and show that each term is non-negative, or find a way to rewrite the expression as a sum of squares.

Alternatively, we can try to use the AM-GM inequality. However, the terms are fractions, so we need to be careful.

Wait, another idea: Notice that for positive $a, b$, we have $(a^2 - b^2)/(c + d) \geq 0$ if $a \geq b$ and $c + d > 0$. However, we have a cyclic expression, so we need a different approach.

Alternatively, we can try to use the fact that $x^2 - z^2 = (x-z)(x+z)$, and similarly for the other terms. Then we can rewrite the entire expression as:

$(x-z)(x+z)/(y+z) + (y-x)(y+x)/(z+x) + (z-y)(z+y)/(x+y)$

This is still complex, but perhaps we can find a way to show it's non-negative. Let's try to analyze the expression term by term.

If we can show that each term is non-negative, then their sum is non-negative. However, this is not necessarily true. For example, if $x > y > z$, then $(x-z)/(y+z) > 0$, $(y-x)/(z+x) < 0$, and $(z-y)/(x+y) < 0$. So the terms can cancel each other.

Therefore, we need a more sophisticated approach. Let's try to find a common denominator. The common denominator would be $(y+z)(z+x)(x+y)$. Then we need to show that:

$(x-z)(x+z)(x+y) + (y-x)(y+x)(y+z) + (z-y)(z+y)(z+x) \geq 0$

Let's expand each term:

$(x-z)(x+z)(x+y) = (x^2 - z^2)(x+y) = x^3 + x^2y - xz^2 - yz^2$
$(y-x)(y+x)(y+z) = (y^2 - x^2)(y+z) = y^3 + y^2z - yx^2 - x^2z$
$(z-y)(z+y)(z+x) = (z^2 - y^2)(z+x) = z^3 + z^2x - y^2z - y^2x$
Adding these together: $x^3 + x^2y - xz^2 - yz^2 + y^3 + y^2z - yx^2 - x^2z + z^3 + z^2x - y^2z - y^2x$

Combine like terms: $x^3 + y^3 + z^3 + x^2y + y^2z + z^2x - xz^2 - yz^2 - yx^2 - x^2z - y^2z - y^2x$

This is a symmetric expression in $x, y, z$? Let's check:

$x^3 + y^3 + z^3$ is symmetric
$x^2y + y^2z + z^2x$ is symmetric
$-xz^2 - yz^2 - yx^2 - x^2z - y^2z - y^2x$ is symmetric
So the entire expression is symmetric in $x, y, z$. This suggests that we might be able to factor it or show it's non-negative using symmetry.

In fact, this expression can be rewritten as: $(x^3 + y^3 + z^3) + (x^2y + y^2z + z^2x) - (xz^2 + yz^2 + yx^2 + x^2z + y^2z + y^2x)$

This is still complex, but perhaps we can find a way to factor it. Let's try to rearrange terms:

$x^3 + y^3 + z^3 + x^2y + y^2z + z^2x - xz^2 - yz^2 - yx^2 - x^2z - y^2z - y^2x$

Notice that $x^3 + y^3 + z^3 - 3xyz = (x+y+z)(x^2 + y^2 + z^2 - xy - yz - zx)$. However, our expression is different.

Alternatively, we can try to write it as a sum of squares. For positive reals, sums of squares are non-negative.

Alternatively, we can use the AM-GM inequality. However, the terms are not directly amenable to AM-GM.

Wait, let's try another approach. Notice that we can rewrite the original expression as:

$\frac{x^2 - z^2}{y + z} + \frac{y^2 - x^2}{z + x} + \frac{z^2 - y^2}{x + y}$

This is a sum of terms of the form $\frac{a^2 - b^2}{c + d}$. Let's try to manipulate the expression algebraically.

Alternatively, consider the substitution $a = x+y$, $b = y+z$, $c = z+x$. Then we can rewrite the expression in terms of $a, b, c$.

However, this might not simplify things.

Alternatively, notice that $x^2 - z^2 = (x-z)(x+z)$, and similarly for the other terms. Then we can rewrite:

$\frac{(x-z)(x+z)}{y+z} + \frac{(y-x)(y+x)}{z+x} + \frac{(z-y)(z+y)}{x+y}$

This is still complex, but perhaps we can find a way to show it's non-negative.

Wait, let's try to compute the expression in Lean. We'll need to work with real numbers and prove the inequality directly.
<sketch>
import Mathlib

theorem test_theorem (x y z : ℝ) (hx : 0 < x) (hy : 0 < y) (hz : 0 < z) :
  (x^2 - z^2) / (y + z) + (y^2 - x^2) / (z + x) + (z^2 - y^2) / (x + y) ≥ 0 := by
</sketch>
<REPL>
{
  "messages": [
    {
      "severity": "error",
      "pos": {"line": 7, "column": 0},
      "endPos": null,
      "data": "unexpected end of input; expected '{'"
    },
    {
      "severity": "error",
      "pos": {"line": 5, "column": 79},
      "endPos": {"line": 5, "column": 81},
      "data": "unsolved goals\nx y z : ℝ\nhx : 0 < x\nhy : 0 < y\nhz : 0 < z\n⊢ (x ^ 2 - z ^ 2) / (y + z) + (y ^ 2 - x ^ 2) / (z + x) + (z ^ 2 - y ^ 2) / (x + y) ≥ 0"
    }
  ]
}
</REPL>
I'll review the REPL output to identify the issues and refine my approach.

The error message states that there's an "unexpected end of input; expected '{'" at line 7, column 0. This suggests there's a syntax error in my Lean code. Let me check the structure of the theorem statement.

Looking at the theorem statement, it seems like the issue might be with the way I've structured the proof. Let me try a different approach and use Lean's built-in tactics to handle the algebraic manipulation.
<sketch>
import Mathlib

theorem test_theorem (x y z : ℝ) (hx : 0 < x) (hy : 0 < y) (hz : 0 < z) :
  (x^2 - z^2) / (y + z) + (y^2 - x^2) / (z + x) + (z^2 - y^2) / (x + y) ≥ 0 := by
  -- We'll use the fact that this expression can be rewritten as a sum of squares
  -- First, let's clear denominators by multiplying through by (y+z)(z+x)(x+y)
  have h1 : (y + z) > 0 := add_pos hy hz
  have h2 : (z + x) > 0 := add_pos hz hx
  have h3 : (x + y) > 0 := add_pos hx hy

  -- Multiply through by (y+z)(z+x)(x+y) to clear denominators
  have h4 : (x^2 - z^2) / (y + z) + (y^2 - x^2) / (z + x) + (z^2 - y^2) / (x + y) = 
    ((x^2 - z^2) * (z + x) * (x + y) + (y^2 - x^2) * (y + z) * (x + y) + (z^2 - y^2) * (y + z) * (z + x)) / 
    ((y + z) * (z + x) * (x + y)) := by
    field_simp [h1.ne.symm, h2.ne.symm, h3.ne.symm]
    ring

  rw [h4]

  -- Now we need to show that the numerator is non-negative
  apply div_nonneg
  · -- Show numerator is non-negative
    ring_nf
    nlinarith [sq_nonneg (x - y), sq_nonneg (y - z), sq_nonneg (z - x),
              sq_nonneg (x - y + z), sq_nonneg (x + y - z), sq_nonneg (x + y + z)]
  · -- Show denominator is positive
    apply mul_pos
    apply mul_pos
    all_goals linarith
</sketch>
<REPL>
{
  "messages": [
    {
      "severity": "error",
      "pos": {"line": 25, "column": 4},
      "endPos": {"line": 26, "column": 82},
      "data": "linarith failed to find a contradiction\ncase ha.a\nx y z : ℝ\nhx : 0 < x\nhy : 0 < y\nhz : 0 < z\nh1 : y + z > 0\nh2 : z + x > 0\nh3 : x + y > 0\nh4 :\n  (x ^ 2 - z ^ 2) / (y + z) + (y ^ 2 - x ^ 2) / (z + x) + (z ^ 2 - y ^ 2) / (x + y) =\n    ((x ^ 2 - z ^ 2) * (z + x) * (x + y) + (y ^ 2 - x ^ 2) * (y + z) * (x + y) + (z ^ 2 - y ^ 2) * (y + z) * (z + x)) /\n      ((y + z) * (z + x) * (x + y))\na✝ : 0 > -(x ^ 2 * z ^ 2) - x ^ 2 * y ^ 2 + (x ^ 4 - z ^ 2 * y ^ 2) + z ^ 4 + y ^ 4\n⊢ False failed"
    },
    {
      "severity": "error",
      "pos": {"line": 28, "column": 4},
      "endPos": {"line": 28, "column": 17},
      "data": "tactic 'apply' failed, failed to unify\n  0 < ?a * ?b\nwith\n  0 ≤ (y + z) * (z + x) * (x + y)\ncase hb\nx y z : ℝ\nhx : 0 < x\nhy : 0 < y\nhz : 0 < z\nh1 : y + z > 0\nh2 : z + x > 0\nh3 : x + y > 0\nh4 :\n  (x ^ 2 - z ^ 2) / (y + z) + (y ^ 2 - x ^ 2) / (z + x) + (z ^ 2 - y ^ 2) / (x + y) =\n    ((x ^ 2 - z ^ 2) * (z + x) * (x + y) + (y ^ 2 - x ^ 2) * (y + z) * (x + y) + (z ^ 2 - y ^ 2) * (y + z) * (z + x)) /\n      ((y + z) * (z + x) * (x + y))\n⊢ 0 ≤ (y + z) * (z + x) * (x + y)"
    }
  ]
}
</REPL>
Based on what the REPL is showing, I'll recalibrate my solution strategy.

The error message reveals two main issues:

The linarith tactic failed to prove that the numerator is non-negative. The goal after ring_nf is:

0 > -(x ^ 2 * z ^ 2) - x ^ 2 * y ^ 2 + (x ^ 4 - z ^ 2 * y ^ 2) + z ^ 4 + y ^ 4
This suggests that the expression after simplification is not what I expected. Let me check the algebraic manipulation manually.

The second error is about the denominator not being properly shown as non-negative.

Let me reconsider the algebraic approach. The expression after clearing denominators should be: $(x^2 - z^2)(z+x)(x+y) + (y^2 - x^2)(y+z)(x+y) + (z^2 - y^2)(y+z)(z+x)$

Let me expand this manually and see if it can be written as a sum of squares or something easily shown to be non-negative.
<sketch>
import Mathlib

theorem test_theorem (x y z : ℝ) (hx : 0 < x) (hy : 0 < y) (hz : 0 < z) :
  (x^2 - z^2) / (y + z) + (y^2 - x^2) / (z + x) + (z^2 - y^2) / (x + y) ≥ 0 := by
  -- We'll use the fact that this expression can be rewritten as a sum of squares
  -- First, let's clear denominators by multiplying through by (y+z)(z+x)(x+y)
  have h1 : (y + z) > 0 := add_pos hy hz
  have h2 : (z + x) > 0 := add_pos hz hx
  have h3 : (x + y) > 0 := add_pos hx hy

  -- Multiply through by (y+z)(z+x)(x+y) to clear denominators
  have h4 : (x^2 - z^2) / (y + z) + (y^2 - x^2) / (z + x) + (z^2 - y^2) / (x + y) = 
    ((x^2 - z^2) * (z + x) * (x + y) + (y^2 - x^2) * (y + z) * (x + y) + (z^2 - y^2) * (y + z) * (z + x)) / 
    ((y + z) * (z + x) * (x + y)) := by
    field_simp [h1.ne.symm, h2.ne.symm, h3.ne.symm]
    ring

  rw [h4]

  -- Now we need to show that the numerator is non-negative
  apply div_nonneg
  · -- Show numerator is non-negative
    -- Let's expand the numerator and see if it can be written as a sum of squares
    have num_eq : (x^2 - z^2) * (z + x) * (x + y) + (y^2 - x^2) * (y + z) * (x + y) + (z^2 - y^2) * (y + z) * (z + x) = 
      (x^2 - z^2) * (z + x) * (x + y) + (y^2 - x^2) * (y + z) * (x + y) + (z^2 - y^2) * (y + z) * (z + x) := by rfl
    rw [num_eq]
    -- Expand each term
    ring_nf
    -- After ring_nf, let's see what we get
    sorry -- Will continue with the algebraic manipulation
  · -- Show denominator is positive
    apply mul_pos
    apply mul_pos
    all_goals linarith
</sketch>
<REPL>
{
  "sorries": [
    {
      "proofState": 2,
      "pos": {"line": 31, "column": 4},
      "goal": "case ha\nx y z : ℝ\nhx : 0 < x\nhy : 0 < y\nhz : 0 < z\nh1 : y + z > 0\nh2 : z + x > 0\nh3 : x + y > 0\nh4 :\n  (x ^ 2 - z ^ 2) / (y + z) + (y ^ 2 - x ^ 2) / (z + x) + (z ^ 2 - y ^ 2) / (x + y) =\n    ((x ^ 2 - z ^ 2) * (z + x) * (x + y) + (y ^ 2 - x ^ 2) * (y + z) * (x + y) + (z ^ 2 - y ^ 2) * (y + z) * (z + x)) /\n      ((y + z) * (z + x) * (x + y))\nnum_eq :\n  (x ^ 2 - z ^ 2) * (z + x) * (x + y) + (y ^ 2 - x ^ 2) * (y + z) * (x + y) + (z ^ 2 - y ^ 2) * (y + z) * (z + x) =\n    (x ^ 2 - z ^ 2) * (z + x) * (x + y) + (y ^ 2 - x ^ 2) * (y + z) * (x + y) + (z ^ 2 - y ^ 2) * (y + z) * (z + x)\n⊢ 0 ≤ -(x ^ 2 * z ^ 2) - x ^ 2 * y ^ 2 + (x ^ 4 - z ^ 2 * y ^ 2) + z ^ 4 + y ^ 4",
      "endPos": {"line": 31, "column": 9}
    }
  ],
  "messages": [
    {
      "severity": "error",
      "pos": {"line": 33, "column": 4},
      "endPos": {"line": 33, "column": 17},
      "data": "tactic 'apply' failed, failed to unify\n  0 < ?a * ?b\nwith\n  0 ≤ (y + z) * (z + x) * (x + y)\ncase hb\nx y z : ℝ\nhx : 0 < x\nhy : 0 < y\nhz : 0 < z\nh1 : y + z > 0\nh2 : z + x > 0\nh3 : x + y > 0\nh4 :\n  (x ^ 2 - z ^ 2) / (y + z) + (y ^ 2 - x ^ 2) / (z + x) + (z ^ 2 - y ^ 2) / (x + y) =\n    ((x ^ 2 - z ^ 2) * (z + x) * (x + y) + (y ^ 2 - x ^ 2) * (y + z) * (x + y) + (z ^ 2 - y ^ 2) * (y + z) * (z + x)) /\n      ((y + z) * (z + x) * (x + y))\n⊢ 0 ≤ (y + z) * (z + x) * (x + y)"
    }
  ]
}
</REPL>
Let me study the REPL output to identify issues and correct my approach.

The key insight from the REPL is that after ring_nf, the goal becomes:

⊢ 0 ≤ -(x ^ 2 * z ^ 2) - x ^ 2 * y ^ 2 + (x ^ 4 - z ^ 2 * y ^ 2) + z ^ 4 + y ^ 4
This expression can be rearranged as:

x^4 + y^4 + z^4 - x^2*y^2 - x^2*z^2 - y^2*z^2
This is a well-known expression that can be written as:

(1/2) * ((x^2 - y^2)^2 + (y^2 - z^2)^2 + (z^2 - x^2)^2)
Which is always non-negative since it's a sum of squares.

Let me complete the proof:
<sketch>
import Mathlib

theorem test_theorem (x y z : ℝ) (hx : 0 < x) (hy : 0 < y) (hz : 0 < z) :
  (x^2 - z^2) / (y + z) + (y^2 - x^2) / (z + x) + (z^2 - y^2) / (x + y) ≥ 0 := by
  -- We'll use the fact that this expression can be rewritten as a sum of squares
  -- First, let's clear denominators by multiplying through by (y+z)(z+x)(x+y)
  have h1 : (y + z) > 0 := add_pos hy hz
  have h2 : (z + x) > 0 := add_pos hz hx
  have h3 : (x + y) > 0 := add_pos hx hy

  -- Multiply through by (y+z)(z+x)(x+y) to clear denominators
  have h4 : (x^2 - z^2) / (y + z) + (y^2 - x^2) / (z + x) + (z^2 - y^2) / (x + y) = 
    ((x^2 - z^2) * (z + x) * (x + y) + (y^2 - x^2) * (y + z) * (x + y) + (z^2 - y^2) * (y + z) * (z + x)) / 
    ((y + z) * (z + x) * (x + y)) := by
    field_simp [h1.ne.symm, h2.ne.symm, h3.ne.symm]
    ring

  rw [h4]

  -- Now we need to show that the numerator is non-negative
  apply div_nonneg
  · -- Show numerator is non-negative
    -- The numerator after clearing denominators is:
    -- (x^2 - z^2) * (z + x) * (x + y) + (y^2 - x^2) * (y + z) * (x + y) + (z^2 - y^2) * (y + z) * (z + x)
    -- After ring_nf, this becomes: x^4 + y^4 + z^4 - x^2*y^2 - x^2*z^2 - y^2*z^2
    -- This can be written as (1/2) * ((x^2 - y^2)^2 + (y^2 - z^2)^2 + (z^2 - x^2)^2)
    have num_eq : (x^2 - z^2) * (z + x) * (x + y) + (y^2 - x^2) * (y + z) * (x + y) + (z^2 - y^2) * (y + z) * (z + x) = 
      x^4 + y^4 + z^4 - x^2*y^2 - x^2*z^2 - y^2*z^2 := by ring
    rw [num_eq]
    -- Show that x^4 + y^4 + z^4 - x^2*y^2 - x^2*z^2 - y^2*z^2 ≥ 0
    -- This is equivalent to (1/2) * ((x^2 - y^2)^2 + (y^2 - z^2)^2 + (z^2 - x^2)^2) ≥ 0
    have sqForm : x^4 + y^4 + z^4 - x^2*y^2 - x^2*z^2 - y^2*z^2 = 
      (1/2) * ((x^2 - y^2)^2 + (y^2 - z^2)^2 + (z^2 - x^2)^2) := by ring
    rw [sqForm]
    apply mul_nonneg
    · norm_num
    · apply add_nonneg
      · apply add_nonneg
        · exact sq_nonneg (x^2 - y^2)
        · exact sq_nonneg (y^2 - z^2)
      · exact sq_nonneg (z^2 - x^2)
  · -- Show denominator is positive
    apply mul_pos
    apply mul_pos
    all_goals linarith
</sketch>
<REPL>
{
  "messages": [
    {
      "severity": "error",
      "pos": {"line": 44, "column": 4},
      "endPos": {"line": 44, "column": 17},
      "data": "tactic 'apply' failed, failed to unify\n  0 < ?a * ?b\nwith\n  0 ≤ (y + z) * (z + x) * (x + y)\ncase hb\nx y z : ℝ\nhx : 0 < x\nhy : 0 < y\nhz : 0 < z\nh1 : y + z > 0\nh2 : z + x > 0\nh3 : x + y > 0\nh4 :\n  (x ^ 2 - z ^ 2) / (y + z) + (y ^ 2 - x ^ 2) / (z + x) + (z ^ 2 - y ^ 2) / (x + y) =\n    ((x ^ 2 - z ^ 2) * (z + x) * (x + y) + (y ^ 2 - x ^ 2) * (y + z) * (x + y) + (z ^ 2 - y ^ 2) * (y + z) * (z + x)) /\n      ((y + z) * (z + x) * (x + y))\n⊢ 0 ≤ (y + z) * (z + x) * (x + y)"
    }
  ]
}
</REPL>
The REPL feedback provides valuable insights - let me incorporate these and revise.

The issue is still with the denominator not being shown as positive. The error message indicates that the tactic apply mul_pos is failing to unify with the goal 0 ≤ (y + z) * (z + x) * (x + y). This suggests that the expression (y + z) * (z + x) * (x + y) might not be strictly positive, or the tactic needs to be applied differently.

However, we know that since $x, y, z > 0$, we have $y + z > 0$, $z + x > 0$, and $x + y > 0$. Therefore, their product is strictly positive. Let me fix this:
<sketch>
import Mathlib

theorem test_theorem (x y z : ℝ) (hx : 0 < x) (hy : 0 < y) (hz : 0 < z) :
  (x^2 - z^2) / (y + z) + (y^2 - x^2) / (z + x) + (z^2 - y^2) / (x + y) ≥ 0 := by
  -- We'll use the fact that this expression can be rewritten as a sum of squares
  -- First, let's clear denominators by multiplying through by (y+z)(z+x)(x+y)
  have h1 : (y + z) > 0 := add_pos hy hz
  have h2 : (z + x) > 0 := add_pos hz hx
  have h3 : (x + y) > 0 := add_pos hx hy

  -- Multiply through by (y+z)(z+x)(x+y) to clear denominators
  have h4 : (x^2 - z^2) / (y + z) + (y^2 - x^2) / (z + x) + (z^2 - y^2) / (x + y) = 
    ((x^2 - z^2) * (z + x) * (x + y) + (y^2 - x^2) * (y + z) * (x + y) + (z^2 - y^2) * (y + z) * (z + x)) / 
    ((y + z) * (z + x) * (x + y)) := by
    field_simp [h1.ne.symm, h2.ne.symm, h3.ne.symm]
    ring

  rw [h4]

  -- Now we need to show that the numerator is non-negative
  apply div_nonneg
  · -- Show numerator is non-negative
    -- The numerator after clearing denominators is:
    -- (x^2 - z^2) * (z + x) * (x + y) + (y^2 - x^2) * (y + z) * (x + y) + (z^2 - y^2) * (y + z) * (z + x)
    -- After ring_nf, this becomes: x^4 + y^4 + z^4 - x^2*y^2 - x^2*z^2 - y^2*z^2
    -- This can be written as (1/2) * ((x^2 - y^2)^2 + (y^2 - z^2)^2 + (z^2 - x^2)^2)
    have num_eq : (x^2 - z^2) * (z + x) * (x + y) + (y^2 - x^2) * (y + z) * (x + y) + (z^2 - y^2) * (y + z) * (z + x) = 
      x^4 + y^4 + z^4 - x^2*y^2 - x^2*z^2 - y^2*z^2 := by ring
    rw [num_eq]
    -- Show that x^4 + y^4 + z^4 - x^2*y^2 - x^2*z^2 - y^2*z^2 ≥ 0
    -- This is equivalent to (1/2) * ((x^2 - y^2)^2 + (y^2 - z^2)^2 + (z^2 - x^2)^2) ≥ 0
    have sqForm : x^4 + y^4 + z^4 - x^2*y^2 - x^2*z^2 - y^2*z^2 = 
      (1/2) * ((x^2 - y^2)^2 + (y^2 - z^2)^2 + (z^2 - x^2)^2) := by ring
    rw [sqForm]
    apply mul_nonneg
    · norm_num
    · apply add_nonneg
      · apply add_nonneg
        · exact sq_nonneg (x^2 - y^2)
        · exact sq_nonneg (y^2 - z^2)
      · exact sq_nonneg (z^2 - x^2)
  · -- Show denominator is positive
    -- Since y + z > 0, z + x > 0, x + y > 0, their product is positive
    apply mul_pos
    apply mul_pos
    all_goals linarith
</sketch>
<REPL>
{
  "messages": [
    {
      "severity": "error",
      "pos": {"line": 45, "column": 4},
      "endPos": {"line": 45, "column": 17},
      "data": "tactic 'apply' failed, failed to unify\n  0 < ?a * ?b\nwith\n  0 ≤ (y + z) * (z + x) * (x + y)\ncase hb\nx y z : ℝ\nhx : 0 < x\nhy : 0 < y\nhz : 0 < z\nh1 : y + z > 0\nh2 : z + x > 0\nh3 : x + y > 0\nh4 :\n  (x ^ 2 - z ^ 2) / (y + z) + (y ^ 2 - x ^ 2) / (z + x) + (z ^ 2 - y ^ 2) / (x + y) =\n    ((x ^ 2 - z ^ 2) * (z + x) * (x + y) + (y ^ 2 - x ^ 2) * (y + z) * (x + y) + (z ^ 2 - y ^ 2) * (y + z) * (z + x)) /\n      ((y + z) * (z + x) * (x + y))\n⊢ 0 ≤ (y + z) * (z + x) * (x + y)"
    }
  ]
}
</REPL>
I'll carefully consider the REPL output before proposing a revised solution.

The issue is still with the denominator not being shown as non-negative. The error message indicates that apply mul_pos is failing to unify with the goal 0 ≤ (y + z) * (z + x) * (x + y). This suggests that the expression (y + z) * (z + x) * (x + y) might not be strictly positive, or the tactic needs to be applied differently.

However, we know that since $x, y, z > 0$, we have $y + z > 0$, $z + x > 0$, and $x + y > 0$. Therefore, their product is strictly positive. Let me try a different approach to show this:
<sketch>
import Mathlib

theorem test_theorem (x y z : ℝ) (hx : 0 < x) (hy : 0 < y) (hz : 0 < z) :
  (x^2 - z^2) / (y + z) + (y^2 - x^2) / (z + x) + (z^2 - y^2) / (x + y) ≥ 0 := by
  -- We'll use the fact that this expression can be rewritten as a sum of squares
  -- First, let's clear denominators by multiplying through by (y+z)(z+x)(x+y)
  have h1 : (y + z) > 0 := add_pos hy hz
  have h2 : (z + x) > 0 := add_pos hz hx
  have h3 : (x + y) > 0 := add_pos hx hy

  -- Multiply through by (y+z)(z+x)(x+y) to clear denominators
  have h4 : (x^2 - z^2) / (y + z) + (y^2 - x^2) / (z + x) + (z^2 - y^2) / (x + y) = 
    ((x^2 - z^2) * (z + x) * (x + y) + (y^2 - x^2) * (y + z) * (x + y) + (z^2 - y^2) * (y + z) * (z + x)) / 
    ((y + z) * (z + x) * (x + y)) := by
    field_simp [h1.ne.symm, h2.ne.symm, h3.ne.symm]
    ring

  rw [h4]

  -- Now we need to show that the numerator is non-negative
  apply div_nonneg
  · -- Show numerator is non-negative
    -- The numerator after clearing denominators is:
    -- (x^2 - z^2) * (z + x) * (x + y) + (y^2 - x^2) * (y + z) * (x + y) + (z^2 - y^2) * (y + z) * (z + x)
    -- After ring_nf, this becomes: x^4 + y^4 + z^4 - x^2*y^2 - x^2*z^2 - y^2*z^2
    -- This can be written as (1/2) * ((x^2 - y^2)^2 + (y^2 - z^2)^2 + (z^2 - x^2)^2)
    have num_eq : (x^2 - z^2) * (z + x) * (x + y) + (y^2 - x^2) * (y + z) * (x + y) + (z^2 - y^2) * (y + z) * (z + x) = 
      x^4 + y^4 + z^4 - x^2*y^2 - x^2*z^2 - y^2*z^2 := by ring
    rw [num_eq]
    -- Show that x^4 + y^4 + z^4 - x^2*y^2 - x^2*z^2 - y^2*z^2 ≥ 0
    -- This is equivalent to (1/2) * ((x^2 - y^2)^2 + (y^2 - z^2)^2 + (z^2 - x^2)^2) ≥ 0
    have sqForm : x^4 + y^4 + z^4 - x^2*y^2 - x^2*z^2 - y^2*z^2 = 
      (1/2) * ((x^2 - y^2)^2 + (y^2 - z^2)^2 + (z^2 - x^2)^2) := by ring
    rw [sqForm]
    apply mul_nonneg
    · norm_num
    · apply add_nonneg
      · apply add_nonneg
        · exact sq_nonneg (x^2 - y^2)
        · exact sq_nonneg (y^2 - z^2)
      · exact sq_nonneg (z^2 - x^2)
  · -- Show denominator is positive
    -- Since y + z > 0, z + x > 0, x + y > 0, their product is positive
    have pos1 : 0 < (y + z) * (z + x) := mul_pos h1 h2
    have pos2 : 0 < (y + z) * (z + x) * (x + y) := mul_pos pos1 h3
    exact le_of_lt pos2
</sketch>
<REPL>
{}
</REPL>
Excellent! The REPL shows no errors, which means the proof is complete and correct. Let me verify the final solution:

</think>

import Mathlib

theorem test_theorem (x y z : ℝ) (hx : 0 < x) (hy : 0 < y) (hz : 0 < z) :
  (x^2 - z^2) / (y + z) + (y^2 - x^2) / (z + x) + (z^2 - y^2) / (x + y) ≥ 0 := by
  -- We'll use the fact that this expression can be rewritten as a sum of squares
  -- First, let's clear denominators by multiplying through by (y+z)(z+x)(x+y)
  have h1 : (y + z) > 0 := add_pos hy hz
  have h2 : (z + x) > 0 := add_pos hz hx
  have h3 : (x + y) > 0 := add_pos hx hy

  -- Multiply through by (y+z)(z+x)(x+y) to clear denominators
  have h4 : (x^2 - z^2) / (y + z) + (y^2 - x^2) / (z + x) + (z^2 - y^2) / (x + y) = 
    ((x^2 - z^2) * (z + x) * (x + y) + (y^2 - x^2) * (y + z) * (x + y) + (z^2 - y^2) * (y + z) * (z + x)) / 
    ((y + z) * (z + x) * (x + y)) := by
    field_simp [h1.ne.symm, h2.ne.symm, h3.ne.symm]
    ring

  rw [h4]

  -- Now we need to show that the numerator is non-negative
  apply div_nonneg
  · -- Show numerator is non-negative
    -- The numerator after clearing denominators is:
    -- (x^2 - z^2) * (z + x) * (x + y) + (y^2 - x^2) * (y + z) * (x + y) + (z^2 - y^2) * (y + z) * (z + x)
    -- After ring_nf, this becomes: x^4 + y^4 + z^4 - x^2*y^2 - x^2*z^2 - y^2*z^2
    -- This can be written as (1/2) * ((x^2 - y^2)^2 + (y^2 - z^2)^2 + (z^2 - x^2)^2)
    have num_eq : (x^2 - z^2) * (z + x) * (x + y) + (y^2 - x^2) * (y + z) * (x + y) + (z^2 - y^2) * (y + z) * (z + x) = 
      x^4 + y^4 + z^4 - x^2*y^2 - x^2*z^2 - y^2*z^2 := by ring
    rw [num_eq]
    -- Show that x^4 + y^4 + z^4 - x^2*y^2 - x^2*z^2 - y^2*z^2 ≥ 0
    -- This is equivalent to (1/2) * ((x^2 - y^2)^2 + (y^2 - z^2)^2 + (z^2 - x^2)^2) ≥ 0
    have sqForm : x^4 + y^4 + z^4 - x^2*y^2 - x^2*z^2 - y^2*z^2 = 
      (1/2) * ((x^2 - y^2)^2 + (y^2 - z^2)^2 + (z^2 - x^2)^2) := by ring
    rw [sqForm]
    apply mul_nonneg
    · norm_num
    · apply add_nonneg
      · apply add_nonneg
        · exact sq_nonneg (x^2 - y^2)
        · exact sq_nonneg (y^2 - z^2)
      · exact sq_nonneg (z^2 - x^2)
  · -- Show denominator is positive
    -- Since y + z > 0, z + x > 0, x + y > 0, their product is positive
    have pos1 : 0 < (y + z) * (z + x) := mul_pos h1 h2
    have pos2 : 0 < (y + z) * (z + x) * (x + y) := mul_pos pos1 h3
    exact le_of_lt pos2
\end{lstlisting}

\textbf{Analysis of the output.} StepFun-Prover initially failed to solve the problem in several attempts prior to line 263. Through insightful analysis of REPL feedback (lines 243-262), the model identified the critical mathematical structure - a sum of perfect squares - underlying the problem. This recognition enabled the prover to correct its previous approach, derive the appropriate proof strategy, and ultimately generate a complete, correct solution.

\newpage

\end{document}